\documentclass[11pt,letterpaper]{article}
\usepackage{authblk}
\usepackage{emnlp2017}
\usepackage{times}
\usepackage{latexsym}
\usepackage{url}
\usepackage{graphicx}
\usepackage{booktabs}
\usepackage{todonotes}

\emnlpfinalcopy

\def\emnlppaperid{***}

\title{Unsupervised, Knowledge-Free, and Interpretable \\ Word Sense Disambiguation}

\author[$\ddag$]{\bf Alexander Panchenko}
\author[$\ddag$]{\bf Fide Marten}
\author[$\ddag$]{\bf Eugen Ruppert}
\author[$\dag$]{\bf Stefano Faralli}
\author[$*$]{\\\bf Dmitry Ustalov} 
\author[$\dag$]{\bf Simone Paolo Ponzetto}
\author[$\ddag$]{\bf Chris Biemann}

\affil[$\ddag$]{Language Technology Group, Department of Informatics, Universit\"{a}t Hamburg, Germany}
\affil[$\dag$]{Web and Data Science Group, Department of Informatics, Universit\"{a}t Mannheim, Germany}
\affil[$*$]{Institute of Natural Sciences and Mathematics, Ural Federal University, Russia} % @Dima: we will include both of your institutions in the final version, here one for the sake of space
\affil[ ]{\href{mailto:panchenko@informatik.uni-hamburg.de}{\{panchenko,marten,ruppert,biemann\}@informatik.uni-hamburg.de}}
\affil[ ]{\href{mailto:simone@informatik.uni
-mannheim.de}{\{simone,stefano\}@informatik.uni-mannheim.de}}
\affil[ ]{\href{mailto:dmitry.ustalov@urfu.ru}{dmitry.ustalov@urfu.ru}}

\date{}

\begin{document}

\maketitle

\begin{abstract}
Interpretability of a predictive model is a powerful feature that gains the trust of users in the correctness of the predictions.  In word sense disambiguation (WSD), \textit{knowledge-based} systems tend to be much more interpretable than \textit{knowledge-free} counterparts as they rely on the wealth of manually-encoded elements representing word senses, such as hypernyms, usage examples, and images. We present a WSD system that bridges the gap between these two so far disconnected groups of methods. Namely, our system, providing access to several state-of-the-art WSD models, aims to be \textit{interpretable} as a knowledge-based system while it remains completely \textit{unsupervised} and \textit{knowledge-free}. The presented tool features a Web interface for all-word disambiguation of texts that makes the sense predictions human readable by providing interpretable word sense inventories, sense representations, and disambiguation results. We provide a public API, enabling seamless integration. 
\end{abstract}

\section{Introduction}

The notion of word sense is central to computational lexical semantics. Word senses can be either \textit{encoded manually} in lexical resources or \textit{induced automatically} from text. The former knowledge-based sense representations, such as those found in the BabelNet lexical semantic network~\cite{Navigli:12}, are easily interpretable by humans due to the presence of definitions, usage examples, taxonomic relations, related words, and images. The cost of such interpretability is that every element mentioned above is encoded manually in one of the underlying resources, such as Wikipedia. Unsupervised knowledge-free approaches, e.g.~\cite{DiMarco:13,Bartunov:16},
% for final:Hope2013a,
require no manual labor, but the resulting sense representations lack the above-mentioned features enabling interpretability. For instance, systems based on sense embeddings are based on dense uninterpretable vectors. Therefore, the meaning of a sense can be interpreted only on the basis of a list of related senses.    

We present a system that brings interpretability of the knowledge-based sense representations into the world of unsupervised knowledge-free WSD models. The contribution of this paper is the first \textit{system} for word sense induction and disambiguation, which is unsupervised, knowledge-free, and interpretable at the same time. The system is based on the WSD approach of~\newcite{Panchenko:17} and is designed to reach interpretability level of knowledge-based systems, such as Babelfy~\cite{Moro:14}, within an unsupervised knowledge-free framework. Implementation of the system is open source.\footnote{\url{https://github.com/uhh-lt/wsd}} A live demo featuring several disambiguation models is available online.\footnote{\url{http://jobimtext.org/wsd}}

\begin{figure*}
\begin{center}
\includegraphics[width=1.0\textwidth]{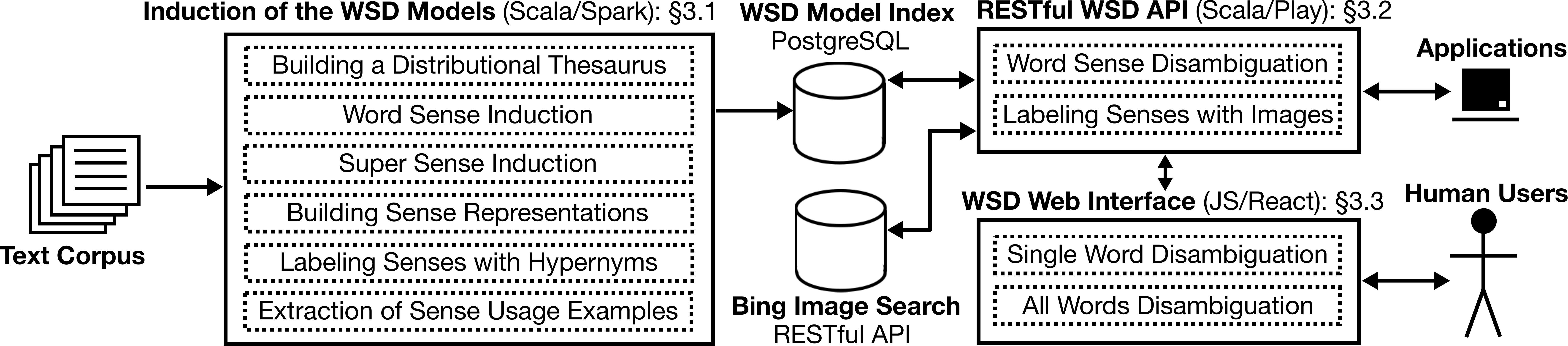}
\end{center}
\caption{Software and functional architecture of the WSD system. }
\label{fig:arch}
\end{figure*}

\section{Related Work}

In this section, we list prominent WSD systems with openly available implementations. 

\paragraph{Knowledge-Based and/or Supervised Systems}

IMS~\cite{Zhong:10} is a supervised all-words WSD system that allows users to integrate additional features and different classifiers. By default, the system relies on the linear support vector machines with multiple features. The AutoExtend~\cite{Rothe:15} approach can be used to learn embeddings for lexemes and synsets of a lexical resource. These representations were successfully used to perform WSD using the IMS.   

DKPro WSD~\cite{Miller:13} is a modular, extensible Java framework for word sense disambiguation. It implements multiple WSD methods and also provides an interface to evaluation datasets. PyWSD\footnote{\url{https://github.com/alvations/pywsd}} project also provides implementations of popular WSD methods, but these are implemented in the Python language.

Babelfy~\cite{Moro:14} is a system based on the BabelNet that implements a multilingual graph-based approach to entity linking and WSD based on the identification of candidate meanings using the densest subgraph heuristic.

\paragraph{Knowledge-Free and Unsupervised Systems} % \todo{These both are WSI. Are you sure about WSD?} The original papers use the WSD benchmarks, the systems include WSD functionality (go eg to the adagram github readme and check it).

\newcite{Neelakantan:14} proposed a multi-sense extension of the Skip-gram model that features an open implementation. AdaGram \cite{Bartunov:16} is a system that learns sense embeddings using a Bayesian extension of the Skip-gram model and provides WSD functionality based on the induced sense inventory. SenseGram \cite{Pelevina:16} is a system that transforms word embeddings to sense embeddings via graph clustering and uses them for WSD. Other methods to learn sense embeddings %, such as \shortcite{li2015multi} 
were proposed, but these do not feature open implementations for WSD. 

Among all listed systems, only Babelfy implements a user interface supporting interpretable visualization of the disambiguation results.

\section{Unsupervised Knowledge-Free Interpretable WSD}

This section describes (1) how WSD models are learned in an unsupervised way from text and (2) how the system uses these models to enable human interpretable disambiguation in context.

\subsection{Induction of the WSD Models}

Figure~\ref{fig:arch} presents architecture of the WSD system. As one may observe, no human labor is used to learn interpretable sense representations and the corresponding disambiguation models. Instead, these are induced from the input text corpus using the JoBimText approach~\cite{Biemann:13} implemented using the Apache Spark framework\footnote{\url{http://spark.apache.org}}, enabling seamless processing of large text collections. Induction of a WSD model consists of several steps. First, a graph of semantically related words, i.e. a distributional thesaurus, is extracted. Second, word senses are induced by clustering of an ego-network of related words~\cite{Biemann:06}. Each discovered word sense is represented as a cluster of words. Next, the induced sense inventory is used as a pivot to generate sense representations by aggregation of the context clues of cluster words. %for final: as described in \cite{panchenkonoun}.
To improve interpretability of the sense clusters they are labeled with hypernyms, which are in turn extracted from the input corpus using~\newcite{Hearst:92} patterns. Finally, the obtained WSD model is used to retrieve a list of sentences that characterize each sense. Sentences that mention a given word are disambiguated and then ranked by prediction confidence. Top sentences are used as sense usage examples. For more details about the model induction process refer to~\cite{Panchenko:17}. Currently, the following WSD models induced from a text corpus are available: % for final: (selectable from the drop-down list, cf. Figure~\ref{fig:single} (C)):

\textbf{Word senses based on cluster word features}. This model uses the cluster words from the induced word sense inventory as sparse features that represent the sense. 

\textbf{Word senses based on context word features}. This representation is based on a sum of word vectors of all cluster words in the induced sense inventory weighted by distributional similarity scores. 

\textbf{Super senses based on cluster word features}. To build this model, induced word senses are first globally clustered using the Chinese Whispers graph clustering algorithm~\cite{Biemann:06}. The edges in this sense graph are established by disambiguation of the related words \cite{Faralli:16,Ustalov:17}. The resulting clusters represent semantic classes grouping words sharing a common hypernym, e.g. ``animal''. This set of semantic classes is used as an automatically learned inventory of super senses: There is only one global sense inventory shared among all words in contrast to the two previous traditional ``per word'' models. Each semantic class is labeled with hypernyms. This model uses words belonging to the semantic class as features. 

\textbf{Super senses based on context word features}. This model relies on the same semantic classes as the previous one but, instead, sense representations are obtained by averaging vectors of words sharing the same class.

\subsection{WSD API} 
To enable fast access to the sense inventories and effective parallel predictions, the WSD models obtained at the previous step were indexed in a relational database.\footnote{\url{https://www.postgresql.org}} In particular, each word sense is represented by its hypernyms, related words, and  usage examples. Besides, for each sense, the database stores an aggregated context word representation in the form of a serialized object containing a sparse vector in the Breeze format.\footnote{\url{https://github.com/scalanlp/breeze}} During the disambiguation phrase, the input context is represented in the same sparse feature space and the classification is reduced to the computation of the cosine similarity between the context vector and the vectors of the candidate senses retrieved from the database. This back-end is implemented as a RESTful API using the Play framework.\footnote{\url{https://www.playframework.com}}  

\subsection{User Interface for Interpretable WSD}

\begin{figure*}
\begin{center}
\includegraphics[width=1.0\textwidth]{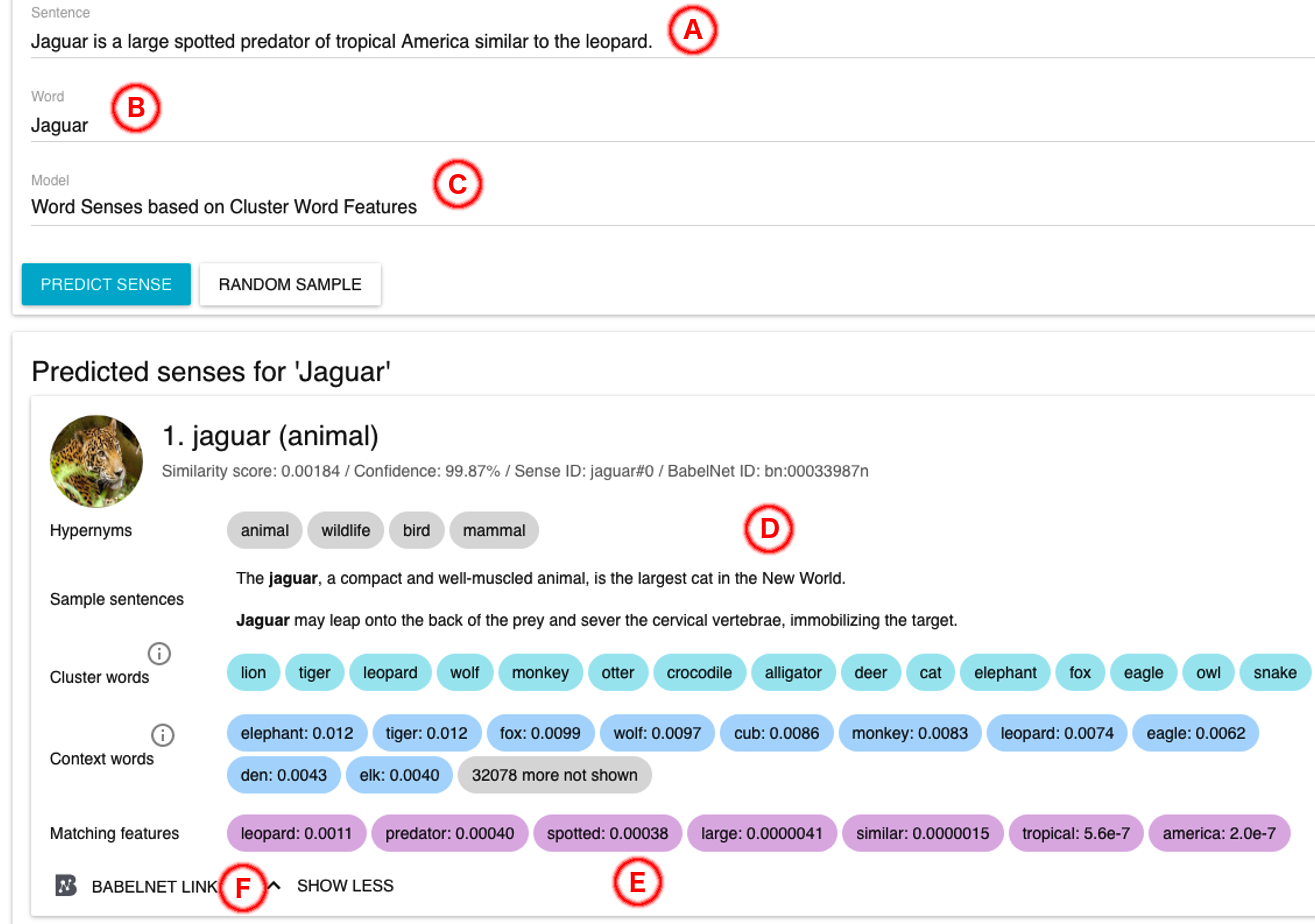} 
\end{center}
\caption{Single word disambiguation mode: results of disambiguation of the word ``Jaguar'' (B) in the sentence ``\textit{Jaguar} is a large spotted predator of tropical America similar to the leopard.'' (A) using the WSD disambiguation model based on cluster word features (C). The predicted sense is summarized with a hypernym and an image (D) and further represented with usage examples, semantically related words, and typical context clues. Each of these elements is extracted automatically. The reasons of the predictions are provided in terms of common sparse features of the input sentence and a sense representation (E).  The induced senses are linked to BabelNet using the method of \newcite{Faralli:16} (F). 
}
\label{fig:single}
\end{figure*}

The graphical user interface of our system is implemented as a single page Web application using the React framework.\footnote{\url{https://facebook.github.io/react}} The application performs disambiguation of a text entered by a user. In particular, the Web application features two modes: 

\textbf{Single word disambiguation mode} is illustrated in Figure~\ref{fig:single}. In this mode, a user specifies an ambiguous word and its context. The output of the system is a ranked list of all word senses of the ambiguous word ordered by relevance to the input context. By default, only the best matching sense is displayed. The user can quickly understand the meaning of each induced sense by looking at the hypernym and the image representing the sense. \newcite{Faralli:12} showed that Web search engines can be used to acquire information about word senses. We assign an image to each word in the cluster by querying an image search API\footnote{\url{https://azure.microsoft.com/en-us/services/cognitive-services/search}} using a query composed of the ambiguous word and its hypernym, e.g. ``jaguar animal''. The first hit of this query is selected to represent the induced word sense. 
Interpretability of each sense is further ensured by providing to the user the list of related senses, the list of the most salient context clues, and the sense usage examples (cf. Figure~\ref{fig:single}). Note that all these elements are obtained without manual intervention. 

Finally, the system provides the reasons behind the sense predictions by displaying context words triggered the prediction. Each common feature is clickable, so a user is able to trace back sense cluster words containing this context feature. 

\begin{figure*}
\begin{center}
\includegraphics[width=1.05\textwidth]{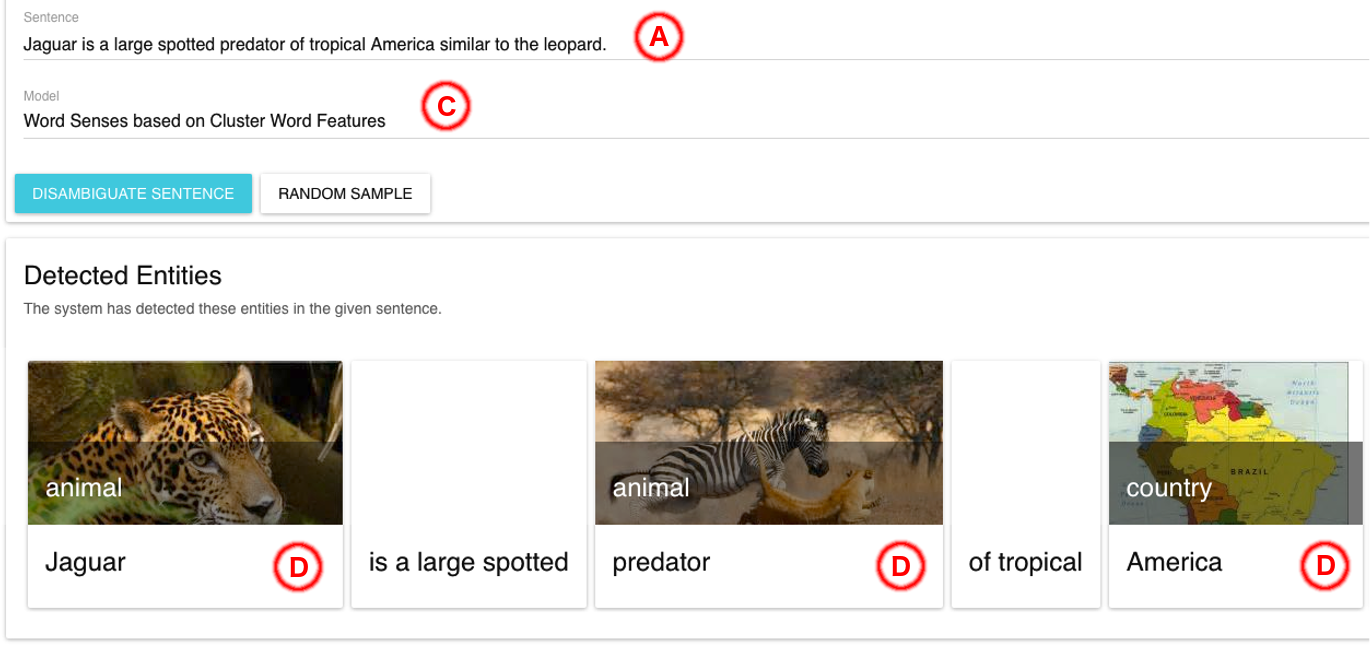} 
\end{center}
\caption{All words disambiguation mode: results of disambiguation of all nouns in a sentence.  }
\label{fig:all}
\end{figure*}

\textbf{All words disambiguation mode} is illustrated in Figure~\ref{fig:all}. In this mode, the system performs disambiguation of all nouns and entities in the input text. First, the text is processed with a part-of-speech and a named entity taggers.\footnote{\url{http://www.scalanlp.org}} Next, each detected noun or entity is disambiguated in the same way as in the single word disambiguation mode described above, yet the disambiguation results are represented as annotations of a running text. The best matching sense is represented by a hypernym and an image as depicted in Figure~\ref{fig:all}. This mode performs ``semantification'' of a text, which can, for instance, assist language learners with the understanding of a text in a foreign language: Meaning of unknown to the learner words can be deduced from hypernyms and images.

\section{Evaluation}

In our prior work~\cite{Panchenko:17}, we performed a thorough evaluation of the method implemented in our system on two datasets showing the state-of-the-art performance of the approach as compared to other unsupervised knowledge-free methods for WSD, including participants of the SemEval 2013 Task 13~\cite{Jurgens:13} and two unsupervised knowledge-free WSD systems based on word sense embeddings \cite{Bartunov:16,Pelevina:16}. These evaluations were based on the ``lexical sample'' setting, where the system is expected to predict a sense identifier of the ambiguous word. 

In this section, we perform an extra evaluation that assesses how well hypernyms of ambiguous words are assigned in context by our system. Namely, the task is to assign a correct hypernym of an ambiguous word, e.g. ``animal'' for the word ``Jaguar'' in the context ``\textit{Jaguar} is a large spotted predator of tropical America''. This task does not depend on a fixed sense inventory and evaluates at the same time WSD performance and the quality of the hypernymy labels of the induced senses. 

\begin{table}
\footnotesize
\centering
\begin{tabular}{llll}
\toprule

% & \multicolumn{2}{c}{\bf Used Features } \\
% & \bf Cluster Feat. & \bf Context Feat. & \bf Sense Inv. \\ \midrule

\bf \# Words & \bf \# Senses & \bf Avg. Polysemy & \bf \# Contexts \\ \midrule
%Full & \\
%Predictable &  \\
863 & 2,708 & 3.13 & 11,712 \\

\bottomrule
\end{tabular}
\caption{Evaluation dataset based on BabelNet. }
\label{tab:dataset}
\end{table}

% 313648

\subsection{Dataset}

In this experiment, we gathered a dataset consisting of definitions of BabelNet 3.7 senses of 1,219 frequent nouns.\footnote{Most of the nouns come from the TWSI~\cite{Biemann:12} dataset, while the remaining nouns were manually selected.} In total, we collected 56,003 sense definitions each labeled with gold hypernyms coming from the IsA relations of BabelNet. 

The average polysemy of words in the gathered dataset was 15.50 senses per word as compared to 2.34 in the induced sense inventory. This huge discrepancy in granularities lead to the fact that some test sentences cannot be correctly predicted by definition: some (mostly rare) BabelNet senses simply have no corresponding sense in the induced inventory.  To eliminate the influence of this idiosyncrasy, we kept only sentences that contain at least one common hypernym with all hypernyms of all induced senses. The statistics of the resulting dataset are presented in Table~\ref{tab:dataset}, it is available in the project repository.

\subsection{Evaluation Metrics}

WSD performance is measured using the accuracy with respect to the sentences labeled with the direct hypernyms (\textit{Hypers}) or an extended set of hypernym including hypernyms of hypernyms (\textit{HyperHypers}). A correct match occurs when the predicted sense has at least one common hypernym with the gold hypernyms of the target word in a test sentence.

\begin{table}
\footnotesize
\centering
\begin{tabular}{llcc}
\toprule

\multicolumn{2}{c}{\bf WSD Model} & \multicolumn{2}{c}{\bf Accuracy}  \\
Inventory & Features & Hypers &  HyperHypers  \\ \midrule

% AdaGram & Random  &  &  \\
%AdaGram & MFS  &  &  \\
%AdaGram & Disambiguation &  &  \\
% \midrule

Word Senses & Random  & 0.257 & 0.610 \\
Word Senses & MFS  & 0.292 & 0.682 \\
Word Senses & Cluster Words & 0.291 & 0.650 \\
Word Senses & Context Words & \underline{\textbf{0.308}} & \underline{\textbf{0.686}} \\ \midrule

Super Senses & Random & 0.001 & 0.001 \\
Super Senses & MFS & 0.001 & 0.001 \\
Super Senses & Cluster Words & \textbf{0.174} & \textbf{0.365} \\
Super Senses & Context Words & 0.086 & 0.188 \\
\bottomrule
\end{tabular}
\caption{Performance of the hypernymy labeling in context on the BabelNet dataset.}
\label{tab:results}
\end{table}

\subsection{Discussion of Results}

\paragraph{Word Senses.} All evaluated models outperform both random and most frequent sense baselines, see Table~\ref{tab:results}. The latter picks the sense that corresponds to the largest sense cluster~\cite{Panchenko:17}. In the case of the traditional ``per word'' inventories, the model based on the context features outperform the models based on cluster words. While sense representations based on the clusters of semantically related words contain highly accurate features, such representations are sparse as one sense contains at most 200 features. As the result, often the model based on the cluster words contain no common features with the features extracted from the input context. The sense representations based on the aggregated context clues are much less sparse, which explains their superior performance.

\paragraph{Super Senses.} In the case of the super sense inventory, the model based solely on the cluster words yielded better results that the context-based model. Note here that (1) the clusters that represent super senses are substantially larger than word sense clusters and thus less sparse, (2) words in the super sense clusters are unweighted in contrast to word sense cluster, thus averaging of word vectors is more noise-prone. Besides, the performance scores of the models based on the super sense inventories are substantially lower compared to their counterparts based on the traditional ``per word'' inventories. Super sense models are able to perform classification for any unknown word missing in the training corpus, but their disambiguation task is more complex (the models need to choose one of 712 classes as compared to an average of 2--3 classes for the ``per word'' inventories). This is illustrated by the near-zero scores of the random and the MFS baselines for this model.

\section{Conclusion}

We present the first openly available word sense disambiguation system that is unsupervised, knowledge-free, and interpretable at the same time. The system performs extraction of word and super sense inventories from a text corpus. The disambiguation models are learned in an unsupervised way for all words in the corpus on the basis on the induced inventories. The user interface of the system provides efficient access to the produced WSD models via a RESTful API or via an interactive Web-based graphical user interface. The system is available online and can be directly used from external applications. The code and the WSD models are open source. Besides, in-house deployments of the system are made easy due to the use of the Docker containers.\footnote{\url{https://www.docker.com}} A prominent direction for future work is supporting more languages and establishing cross-lingual sense links.  

%\todo{Maybe provide a link to a Docker Hub image instead of Docker.com?} we do not have a prebuilt image (just a script instead)

\section*{Acknowledgments}
We acknowledge the support of the DFG under the ``JOIN-T'' project, the RFBR under  project no.~16-37-00354 mol\_a, Amazon via the ``AWS Research Grants'' and Microsoft via the ``Azure for Research'' programs. Finally, we also thank four anonymous reviewers for their helpful comments.

\bibliography{emnlp2017}
\bibliographystyle{emnlp_natbib}

\end{document}